\documentclass[runningheads]{llncs}
\usepackage[T1]{fontenc}
\usepackage{graphicx}
\usepackage[namelimits]{amsmath} 
\usepackage{multirow}
\usepackage{makecell}
\newcommand{\tabincell}[2]{\begin{tabular}{@{}#1@{}}#2\end{tabular}}
\begin{document}
\title{Complex Handwriting Trajectory Recovery: Evaluation Metrics and Algorithm}
\titlerunning{Complex Handwriting Trajectory Recovery}
%
\author{Zhounan Chen\inst{1} \and
Daihui Yang\inst{1} \and
Jinglin Liang\inst{1} \and
Xinwu Liu\inst{4} \and
Yuyi Wang\inst{3,4} \and
Zhenghua Peng\inst{1} \and
Shuangping Huang\thanks{Corresponding author.}\inst{1,2}}
\authorrunning{Z. Chen et al.}
%
\institute{South China University of Technology, Guangzhou, China \and
Pazhou Laboratory, Guangzhou, China \and Swiss Federal Institute of Technology, Zurich, Switzerland \and
CRRC Institute, Zhuzhou, China}
\maketitle              
\begin{abstract}
Many important tasks such as forensic signature verification, calligraphy synthesis, etc, rely on handwriting trajectory recovery of which, however, even an appropriate evaluation metric is still missing. 
Indeed, existing metrics only focus on the writing orders but overlook the fidelity of glyphs. 
Taking both facets into account, we come up with two new metrics, the adaptive intersection on union (AIoU) which eliminates the influence of various stroke widths, and the length-independent dynamic time warping (LDTW) which solves the trajectory-point alignment problem. After that, we then propose a novel handwriting trajectory recovery model named Parsing-and-tracing ENcoder-decoder Network (PEN-Net), in particular for characters with both complex glyph and long trajectory, which was believed very challenging. In the PEN-Net, a carefully designed double-stream parsing encoder parses the glyph structure, and a global tracing decoder overcomes the memory difficulty of long trajectory prediction. 
Our experiments demonstrate that the two new metrics AIoU and LDTW together can truly assess the quality of handwriting trajectory recovery and the proposed PEN-Net exhibits satisfactory performance in various complex-glyph languages including Chinese, Japanese and Indic. The source code is available at https://github.com/ChenZhounan/PEN-Net.

\end{abstract}
\section{Introduction}
Trajectory recovery 
from static handwriting images reveals the natural writing order while ensuring the glyph fidelity. 
There are a lot of applications including forensic signature verification \cite{niels2006automatic,munich2003visual}, calligraphy synthesis and imitation \cite{ISI:000426687100006,zhao2020deep}, handwritten character recognition \cite{rabhi2019handwriting,noubigh2017survey}, handwriting robot \cite{yin2016synthesizing,yao2004trajectory}, etc. 
This paper deals with two main challenges the task of complex handwriting trajectory recovery faces.

\begin{figure}[t]
\begin{center}
\includegraphics[width=0.45\linewidth]{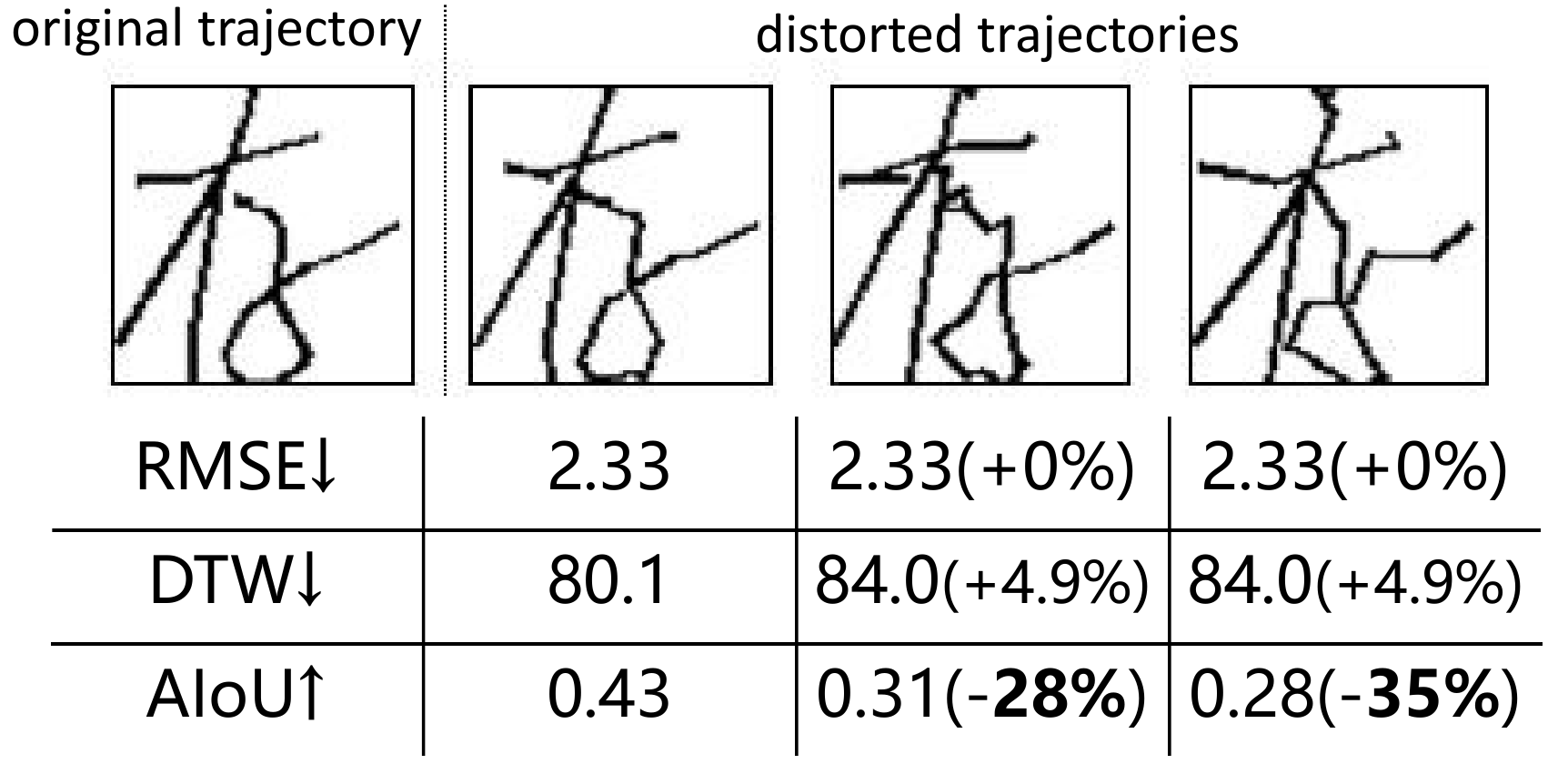}
\end{center}
      \caption{Evaluation scores of distorted trajectories. Three distorted trajectories are obtained by moving half of the points a fixed distance from the original trajectory. The first distorted trajectory shows small glyph structure distortion and its evaluation scores are 2.33, 80.1 and 0.43. The moving angles of points in the other two distorted trajectories are quite different from the first one, hence they show varied degrees of glyph distortion, but this obvious visual differences do not affect the RMSE value and only slightly worsens the DTW value (+4.9 \% and +4.9 \%, respectively). In contrast, the AIoU value exhibits a significant difference (-28 \% and -35 \%, respectively).}
\label{figure1}
\end{figure}

On the one hand, surprisingly, a proper task-targeted evaluation metric for handwriting trajectory recovery is still missing. 
The existing approaches are in three classes, but indeed, they are all with non-negligible drawbacks: 
\begin{enumerate}
    \item Human vision is usually used \cite{pan1991offline,boccignone1993recovering,doermann1995recovery,plamondon1999segmentation,kato2000recovery,nguyen2021online}. This non-quantitative and human-involved method is expensive, time-consuming and inconsistent. 
    \item Some work quantified the recovery quality indirectly through the accuracy of a handwriting recognition model \cite{al2002dynamic,lallican2000off,guo2001forgery}, but the result inevitably depends on the recognition model and hence brings unfairness to evaluation. 
    \item  Direct quantitative metrics have been borrowed from other fields \cite{lau2005directed,nel2008verification,hassaine2013icdar}, without task-targeted adaption. These metrics overlook the glyph fidelity, and some of them even ignore either the image-level fidelity (which is insufficient for our task, compared with glyph fidelity) or the writing order. For example, Fig.\ \ref{figure1} presents the effects of different metrics on glyph fidelity. As it shows, three distorted trajectories show varied degrees of glyph distortion, however, metrics on writing order such as the root mean squared error (RMSE) and the dynamic time warping (DTW) cannot effectively and sensitively reflect their varied degrees of glyph degradation, since trajectory points jitter in the same distance.
    
\end{enumerate}

We propose two evaluation metrics, the adaptive intersection on union (AIoU) and the length-independent dynamic time warping (LDTW). 
AIoU assesses the glyph fidelity and eliminates the influence of various stroke widths. 
LDTW is robust to the number of sequence points and overcomes the evaluation bias of classical DTW \cite{hassaine2013icdar}. 

On the other hand, the existing trajectory recovery algorithms are not good in dealing with characters with both complex glyph and long trajectory, such as Chinese and Japanese scripts, thus we propose a novel learning model named Parsing-and-tracing ENcoder-decoder Network (PEN-Net). 
In the character encoding phase, we add a double-stream parsing encoder in the PEN-Net by creating two branches to analyze stroke context information along two orthogonal dimensions. In the decoding phase, we construct a global tracing decoder to alleviate the drifting problem of long trajectory writing order prediction. 

Our contributions are threefold:
\begin{itemize}
    \item We propose two trajectory recovery evaluation metrics, AIoU to assess glyph correctness which was overlooked by most, if not all, existing quantitative metrics, and LDTW to overcome the evaluation bias of the classical DTW. 
    \item We propose a new trajectory recovery model called PEN-Net by constructing a double-stream parsing encoder and a global tracing decoder to solve the difficulties in the case of complex glyph and long trajectory. 
    \item Our experiments demonstrate that AIoU and LDTW can truly assess the quality of handwritten trajectory recovery and the proposed PEN-Net exhibits satisfactory performance in various complex-glyph datasets including Chinese \cite{liu2011casia}, Japanese \cite{nakagawa2004collection} and Indic \cite{bhunia2018handwriting}.
\end{itemize}

\section{Related Work}

\subsection{Trajectory Recovery Evaluation Metrics}
There is not much work on evaluation metrics for trajectory recovery. Many techniques rely on the human vision to qualitatively assess the recovery quality \cite{pan1991offline,boccignone1993recovering,doermann1995recovery,plamondon1999segmentation,kato2000recovery,nguyen2021online}. 
However, these non-quantitative human evaluations are expensive, time-consuming and inconsistent.

Trajectory recovery has been proved beneficial to handwriting recognition. As a byproduct, one may use the final recognition accuracy to compare the recovery efficacy \cite{al2002dynamic,lallican2000off,guo2001forgery,huang2022agtgan}. Instead of directly assessing the quality of different recovery trajectories, they compare the accuracy of the recognition models among which the only difference is the intermediate recovery trajectory. Though, to some extent, this method reflects the recovery quality, it is usually disturbed by the recognition model, and only provides a relative evaluation. 

Most of direct quantitative metrics only focus on the evaluation of the writing order. 
Lau \textit{et al.}\  \cite{lau2005directed} designed a ranking model to assess the order of stroke endpoints, and Nel \textit{et al.}\  \cite{nel2008verification} designed a hidden Markov model to assess the writing order of local trajectories such as stroke loops. 
However, these methods are unable to assess the sequence points’ deviation to the groundtruth. Hence, these two evaluations are seldom adopted in the subsequent studies.

Metrics borrowed from other fields have also been used, such as the RMSE borrowed from signal processing and the DTW from speech recognition \cite{hassaine2013icdar}. RMSE directly calculates distances between two sequences and strictly limits the number of trajectory points, which makes it hard to use in practice. 
It is too strict to require the recovered trajectory to recall exactly all the groundtruth points one-by-one, since trajectory recovery is an ill-posed problem that the unique recovery solution cannot be obtained without constraints \cite{qiao2006framework}.
DTW introduces an elastic matching mechanism to obtain the most possible alignment between two trajectories. 
However, DTW is not robust to the number of trajectory points, and prefers trajectories with fewer points. Actually, the number of points is irrelevant to the writing order or glyph fidelity, and shouldn't affect the judgement of the recovery quality.

Aforementioned quantitative metrics only focus on the evaluation of the writing order, and neither involves the glyph fidelity. 
Note that the glyph correctness is also an essential aspect of trajectory recovery, since glyphs reflect the content of characters and the writing styles of a specific writer. Only one of the latest trajectory recovery work \cite{archibald2021trace} borrows the metric LPIPS \cite{zhang2018unreasonable} which compares the deep features of images. However, LPIPS is not suitable for such a fine task of trajectory recovery, as we observed that the deep features are not informative enough to distinguish images with only a few pixel-level glyph differences.

\subsection{Trajectory Recovery Algorithms}
Early studies in the 1990s relied on heuristic rules, often including two major modules: local area analysis and global trajectory recovery \cite{boccignone1993recovering,plamondon1999segmentation,jager1996recovering,kato2000recovery,qiao2006framework}. These algorithms are difficult to devise, as they rely on delicate hand-engineered features. Rule-based methods are sophisticated, and not flexible enough to handle most of the practical cases, hence these methods are considered not robust, in particular for characters with both complex glyph and long trajectory. 

Inspired by the remarkable progress in deep-learning-based image analysis and sequence generation over the last few years, deep neural networks are used for trajectory recovery.
Sumi \textit{et al.}\ \cite{sumi2019modality} applied variational autoencoders to mutually convert online trajectories and offline handwritten character images, and their method can handle single-stroke English letters with simple glyph structures. Nevertheless, instead of predicting the entire trajectory from a plain encoding result, we can consider employing a selection mechanism (e.g., attention mechanism) to analyze the glyph structure between the prediction of two successive points, since the relative position of continuous points can be quite variable. 
Zhao \textit{et al.}\  \cite{zhao2018pen,zhao2019drawing} proposed a CNN model to iteratively generate stroke point sequence. However, besides CNNs, we also need to consider applying RNNs to analyze the context in handwriting trajectories, which may contribute to the recovery of long trajectories.

Bhunia \textit{et al.}\ \cite{bhunia2018handwriting} introduced an encoder-decoder model to recover the trajectory of single-stroke Indic scripts. This algorithm employs a one-dimensional feature map to encode characters, however, it needs more spatial information to tackle complex handwritings (on a two-dimensional plane). 
Nguyen \textit{et al.}\ \cite{nguyen2021online} improved the encoder-decoder model by introducing a Gaussian mixture model (GMM), and tried to recover multi-stroke trajectories, in particular Japanese scripts. 
However, since the prediction difficulty of long trajectories remains unsolved, this method does not perform well in the case of complex characters. 
Archibald \textit{et al.}\ \cite{archibald2021trace} adapted the encoder-decoder model to English text with arbitrary width, which attends to text-line-level trajectory recovery, but not designed specifically for complex glyph and long trajectory sequences, either.
\begin{figure*}[t]
\begin{center}
\includegraphics[width=0.8\linewidth]{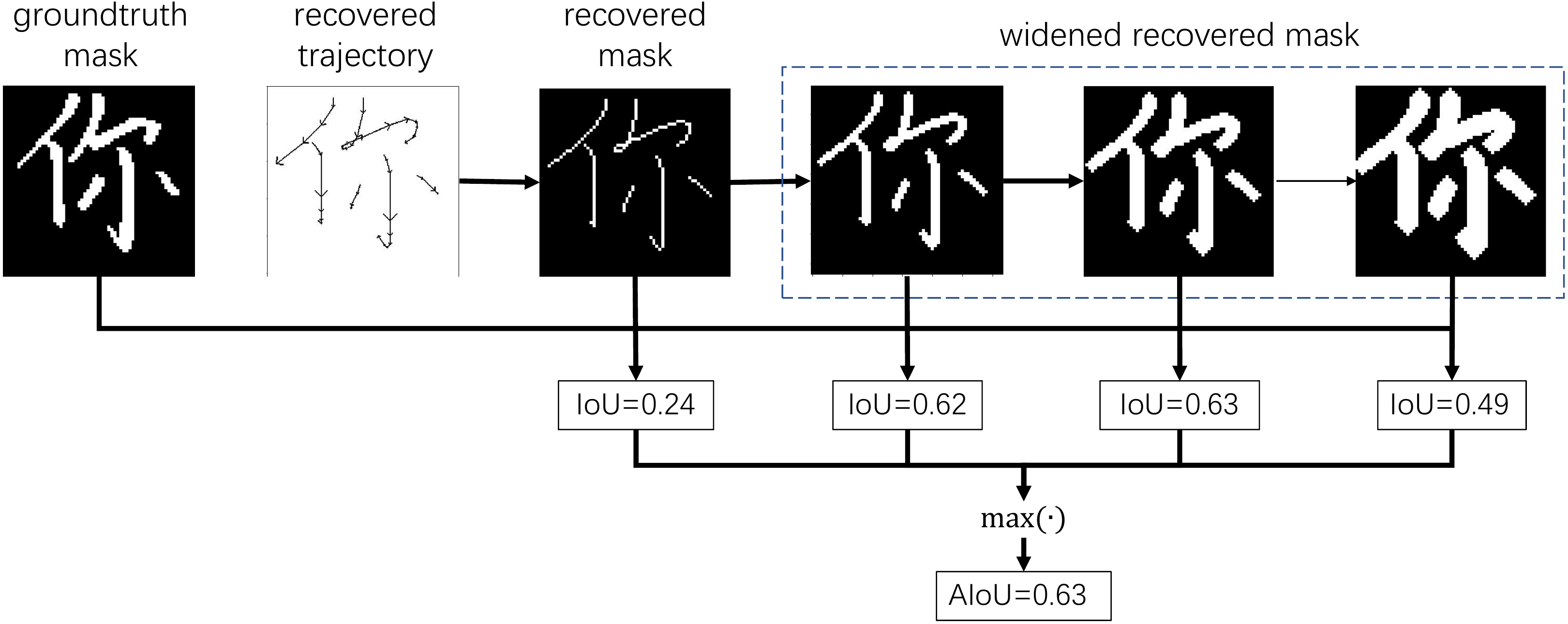}
\end{center}
  \caption{Illustration of the adaptively-dilating mechanism of AIoU.}
\label{figure3-2}
\end{figure*}

\section{Glyph-and-Trajectory Dual-modality Evaluation}
Let $I$ represent a handwritten image, typically in the form of grayscale. The trajectory recovery system takes $I$ as input and predicts the trajectory which can be mathematically represented as a time series $p=\left(p_{1}, \ldots, p_{N}\right)$, where $N$\ is the trajectory length, and $p_{i}=\left(x_{i}, y_{i}, s_{i}^{1}, s_{i}^{2}, s_{i}^{3}\right)$, $x_{i}$ and $y_{i}$ are coordinates of $p_{i}$, $s_{i}^{1}, s_{i}^{2}$ and $s_{i}^{3}$ are pen tip states which are described in detail in Section \ref{DefinitionOfTraj}. And the corresponding groundtruth trajectory is $q=\left(q_{1}, \ldots, q_{M}\right)$ of length $M$. 
\subsection{Adaptive Intersection on Union}
We propose the first glyph fidelity metric, Adaptive Intersection on Union (AIoU). It firstly performs a binarization process on the input image $I$ using a thresholding algorithm, e.g., OTSU \cite{otsu1979threshold}, to obtain the ground-truth binary mask, denoted $G$, which indicates whether a pixel belongs to a character stroke. Meanwhile, the predicted trajectory $p$ is rendered into a bitmap (predicted mask) of width $1$ pixel, by drawing lines between neighboring points if they belong to a stroke, denoted $P$. We define the IoU (Intersection over Union) between $G$ and $P$ as follow, which is similar to the mask IoU \cite{cheng2021boundary} $IoU(G, P)={|G \cap P|}/{|G \cup P|}$.

An input handwritten character usually has various stroke widths while the predicted stroke widths are fixed, nevertheless, the stroke width shouldn't influence the assessment of the glyph similarity. To reduce the impacts of stroke width, we propose a dynamic dilation algorithm to adjust the stroke width adaptively. 
Concretely, as shown in Fig.\  \ref{figure3-2}, we adopt a dilation algorithm \cite{haralick1987image} with a kernel of $3\times 3$ to widen the stroke along until the IoU score reaches the maximum, denoted $AIoU(G, P)$. 
Since the image $I$ is extracted as the binary mask $G$, the ground-truth trajectory of $I$ is not involved in the calculation of the AIoU, making the criteria still effective even without the ground-truth trajectory.

\subsection{Length-independent Dynamic Time Warping}
\label{LDTW}
\begin{figure*}[t]
\begin{center}
\includegraphics[width=0.7\linewidth]{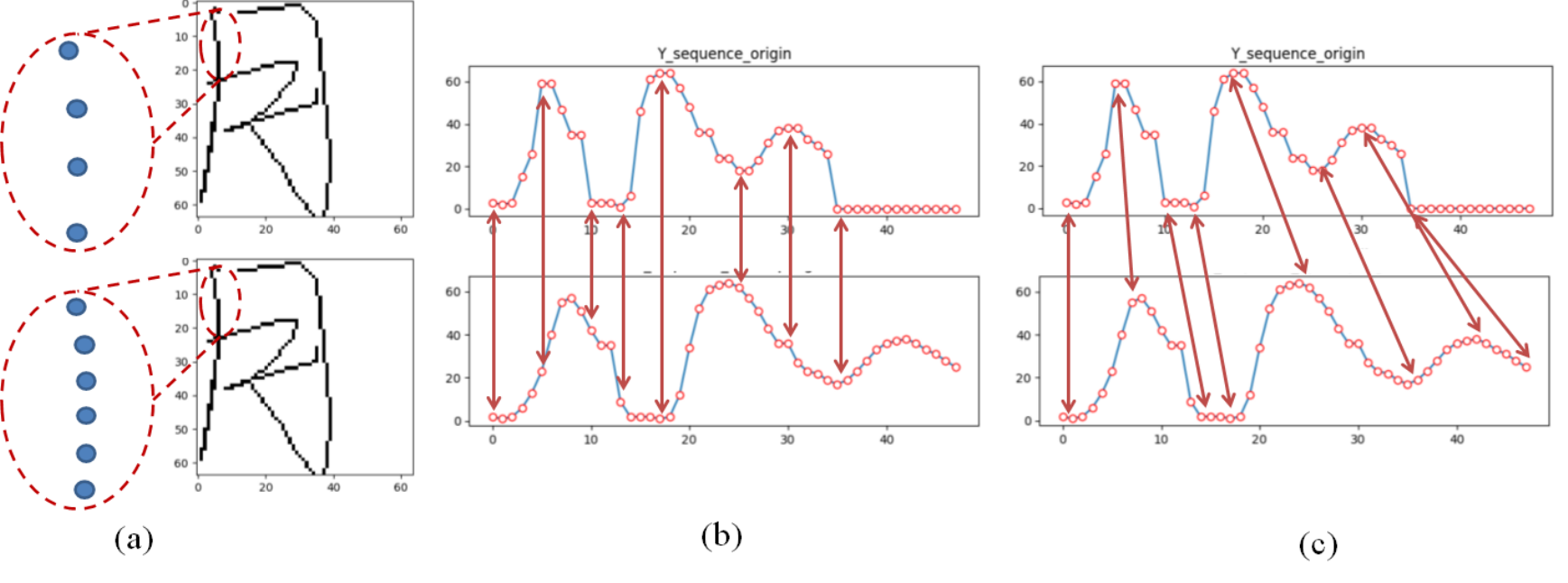}
\end{center}
  \caption{Comparison between the one-to-one and the elastic matching. (a) Original and upsampling handwriting trajectories of a same character. (b) One-to-one, (c) elastic matching of two trajectories. Above and Below waveforms are Y coordinate sequences of the original-sampling and upsampling handwriting trajectory, respectively. Partial correspondence pairs are illustrated as red connection lines.}
\label{figure5-3}
\end{figure*}

Variable lengths make it hard to align handwriting trajectories. As shown in Fig.\ \ref{figure5-3}, when comparing two handwriting trajectories with different lengths, the direct one-to-one stroke-point correspondence cannot represent the correct alignment of strokes. We modify the well-known DTW \cite{hassaine2013icdar} to compare two trajectories whose lengths are allowed to be different, which uses an elastic matching mechanism to obtain the most possible alignment. 

The original DTW relies on the concept of alignment paths, $
DTW(q,p)=\min_{\phi}\left\{\sum_{t=1}^{T} d\left(q_{i_t}, p_{j_t}\right)\right\},
$
where the minimization is taken over all possible alignment paths $\phi$ (which is solved by a dynamic programming), $T \le M+N$ is the alignment length and $d\left(q_{i_t},p_{j_t}\right)$ refers to the (Euclidean) distance between the two potentially matched (determined by $\phi$) points $q_{i_t}$ and $p_{j_t}$.

We observe that the original DTW empirically behaves like a monotonic function of $T$ that is usually proportional to $N$, so it in general prefers short strokes and even gives good score to incomplete strokes, which is what we want to get rid of. Intuitively, this phenomenon is interpretable: DTW is the minimization of a sum of $T$ terms, and $T$ depends on $N$. We suggest a normalized version of DTW, called the length-independent DTW (LDTW)
\begin{equation}
    \label{eq12}
    LDTW(q,p) = \frac{1}{T} DTW(q,p).
\end{equation}

It is worth noting that, since the alignment problem also exists during the training process, we use a soft dynamic time warping loss (SDTW Loss) \cite{cuturi2017soft} to realize a global-alignment optimization, see Section \ref{DefinitionOfTraj}. 

\subsection{Analysis of AIoU and LDTW}
In this part, we firstly investigate how the values of our proposed metrics(AIoU and LDTW) and other recently used metrics change in response to the errors in different magnitudes. Secondly, we analyze the impacts of the changes in the number of trajectory points and stroke width to LDTW and AIoU respectively.

\subsubsection{Error-Sensitivity Analysis}
We simulate a series of common trajectory recovery errors across different magnitude by generating pseudo-predictions with errors such as point or stroke level insertion, deletion, and drift from the ground truth trajectories. Specific implementations of error simulation (e.g., magnitude setting method) is shown in Appendix. 
We conduct the error-sensitivity analysis experiment on the benchmark OLHWDB1.1 (described in section \ref{5.1datasets}), since it contains Chinese characters with complex glyphs and long trajectories. 

We calculate the average score in 1000 randomly-selected sample from OLHWDB1.1 on the metrics of AIoU, LDTW and LPIPS across different error magnitudes. For better visualization, we normalize the values of the three metrics to [0, 1]. As Fig.\ \ref{figure4-8} illustrates, firstly, the values of AIoU and LPIPS, two metrics on the glyph and image level respectively, decrease as the magnitude of the four error types increases. Secondly, the value of LDTW, the proposed quantitative metric for sequence similarity comparison, increases along with the magnitude of the four error types. These two results prove that the three metrics are sensitive to the four errors. Furthermore, as the changing trend of AIoU is faster than LPIPS, the former is more sensitive to the errors than the latter. 

\subsubsection{Invariance Analysis}
In terms of metric invariance, stroke width change and trajectory points number change are two critical factors. The former highlights different handwriting brush strokes (e.g., brushes, pencils, or water pens), which only affects stroke widths and keep the original glyph of the characters. The latter regards to the change of the total number of points in a character to simulate different handwriting speeds. 

\begin{figure}[t]
\begin{center}
\includegraphics[width=0.7\linewidth]{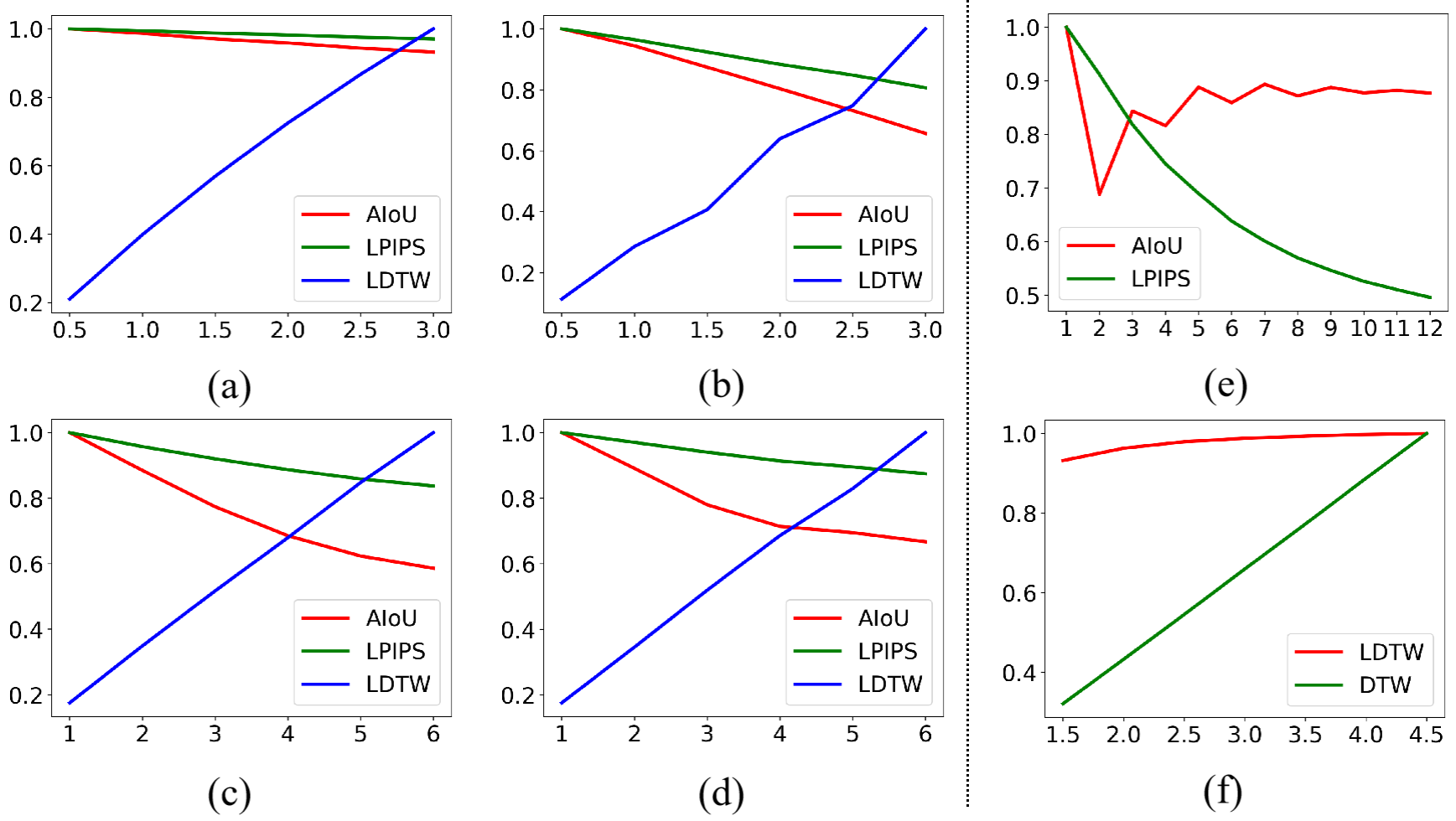}
\end{center}
   \caption{Left: Sensitivity curves across error magnitudes: AIoU, LPIPS\protect\footnotemark[1], LDTW results of 4 error types: (a) Stroke insertion error. (b) Stroke deletion error. (c) Trajectory point drift error. (d) Stroke drift error. X-axes of (a) and (b) are the number of inserted and deleted strokes, respectively. X-axes of (c) and (d) are the drifted pixel distance of point and stroke, respectively. Right: Sensitivity curves across change magnitudes: (e) LPIPS\protect\footnotemark[1] and AIoU results of the change of stroke widths (X-axis). (f) DTW and LDTW results of the change of sample rates (X-axis). Y-axes refer to the normalized metric value for all sub-figures.}
\label{figure4-8}
\end{figure}

This analysis is also based on OLHWDB1.1 and the data preprocessing is the same with the error sensitivity analysis mentioned above. As shown in Fig.\ref{figure4-8}(e), on the overall, the trend of our proposed AIoU is more stable compared with LPIPS as the stroke width of the character increases, indicating that AIoU is more robust to the changes of stroke widths so that it can truly reflect the glyph fidelity of a character. In terms of the trajectory points number change, as shown in Fig.\ref{figure4-8}(f), the value of DTW rises with the increase of the number of points in the character trajectory while our proposed LDTW, on the other hand, shows a smooth and steady trend. This is because LDTW applies length normalization techniques but DTW does not. The results proves that our proposed LDTW is more robust to the changes in the number of points in a character trajectory. 

\section{Parsing-and-tracing ENcoder-decoder Network}
As shown in Fig.\ \ref{figure4-1}, PEN-Net is composed of a double-stream parsing encoder and a global tracing decoder. Taking a static handwriting image as input, the double-stream parsing encoder analyzes the stroke context and parses the glyph structure, obtaining the features that will be used by the global tracing decoder to predict trajectory points. 

\subsection{Double-Stream Parsing Encoder}
\label{Double-Stream Parsing Encoder}
\begin{figure*}[t]
\begin{center}
\includegraphics[width=1\linewidth]{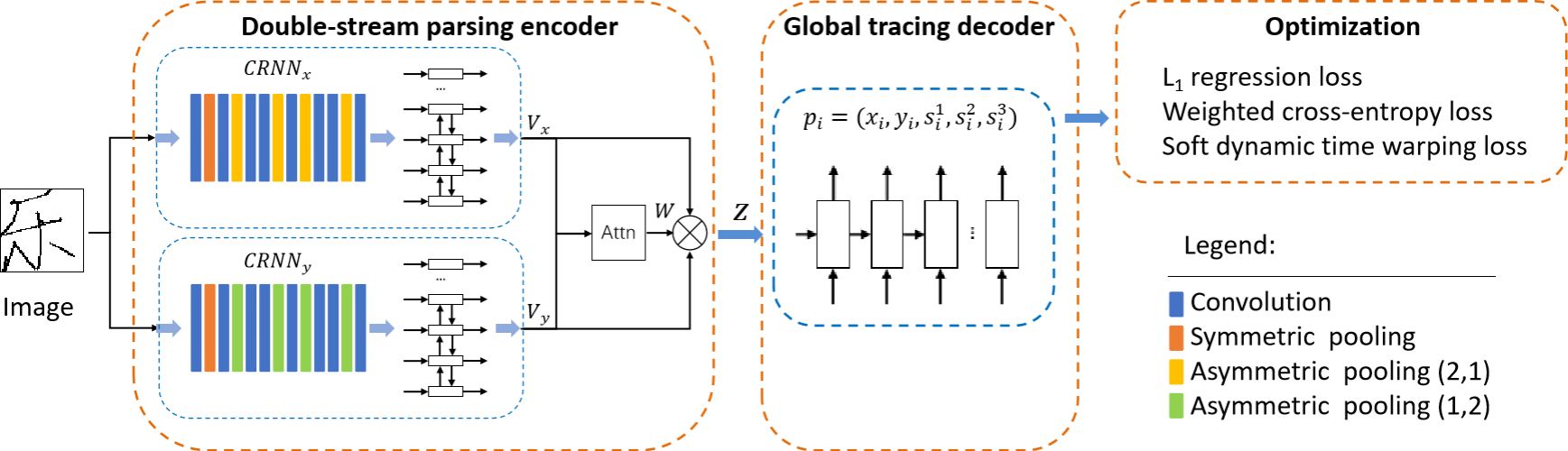}
\end{center}
   \caption{An overview of the Parsing-and-tracing Encoder-decoder Network.}
\label{figure4-1}
\end{figure*}

\begin{figure}[t]
\begin{center}
\includegraphics[width=0.55\linewidth]{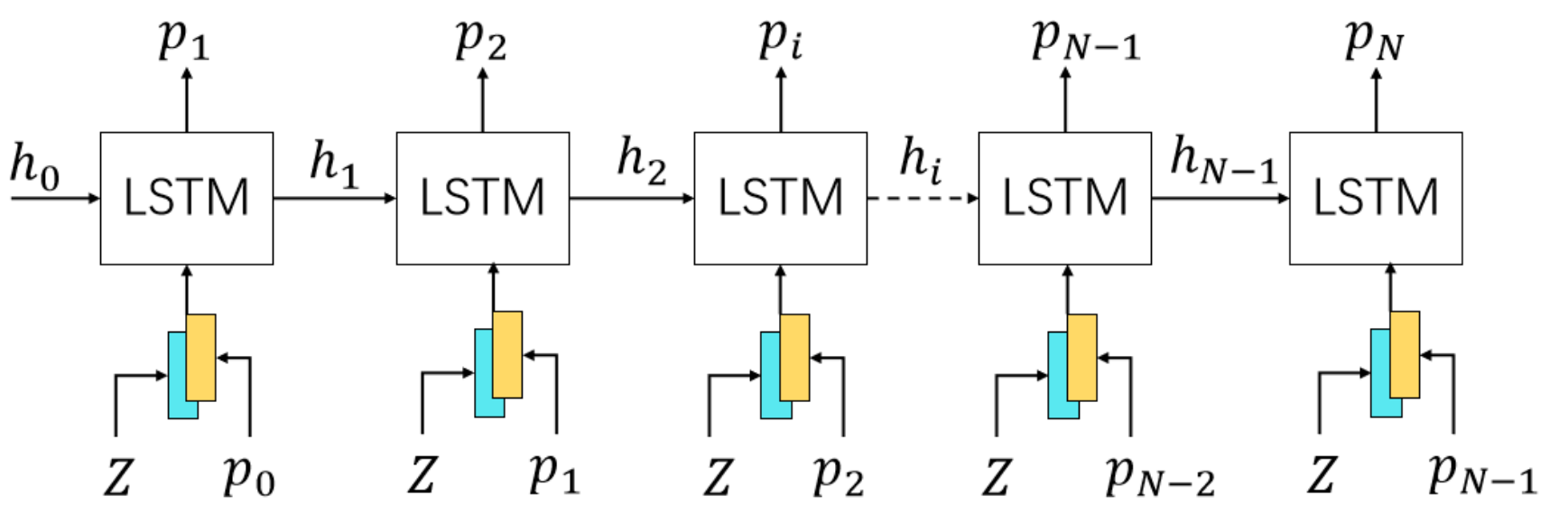}
\end{center}
  \caption{The architecture of global tracing decoder. $Z$ is the glyph parsing feature, $p_{i}$ represent the trajectory point at time $i$, $h_{i}$ is the hidden state of the LSTM at time $i$. $Z$ is concatenated with $p_{i}$. We initialize the hidden state of LSTM decoder with the hidden state outputs of BiLSTMs encoder, which is similar to the \cite{bhunia2018handwriting}.}
\label{figure4-2}
\end{figure}

Existing methods (e.g., DED-Net\cite{bhunia2018handwriting}, Cross-VAE\cite{sumi2019modality}) compress the feature to only one dimension vector, which are not informative enough to maintain the complex two-dimensional information of characters. 
Actually, every two-dimensional stroke can be projected to horizontal and vertical axes. 
In the double-stream parsing encoder, we construct two CRNN \cite{shi2016end} branches denoted as $CRNN_X$ and $CRNN_Y$ to decouple handwriting images to horizontal and vertical features $V_x$ and $V_y$, which are complementary in two perpendicular directions for the parsing of glyph structure. 
Each branch is composed of a CNN to extract the vertical or horizontal stroke features, and a 3-layer BiLSTM to analyze the relationship between strokes, e.g., which stroke should be drawn earlier, what is the relative position between strokes. To extract stroke features of single direction in the CNN of each stream, we use asymmetric poolings, which is found to be effective experimentally. Details of proposed CNNs 
are shown in Fig.\ \ref{figure4-1}.

\footnotetext[1]{Instead of $LPIPS$, we show $1-LPIPS$, for a better visual comparison. }

The stroke region is always sparse in a handwriting image, and the blank background disturbs the stroke feature extraction. To this end, we use an attention mechanism to attend to the stroke foreground. The attention mechanism fuses $V_x$ and $V_y$, and obtains the attention score $s_{i}$ of each feature $v_i$ to let the glyph parsing feature $Z$ focus on the stroke foreground: 
\begin{equation}
\label{eq20}
s_{i}=f\left(v_{i}\right)=U v_{i},
\end{equation}
\begin{equation}
\label{eq21}
w_{i}=\frac{e^{s_{i}}}{\sum_{j=1}^{|V|} e^{s_{j}}},
\end{equation}
\begin{equation}
\label{eq22}
Z=\sum_{i=1}^{|V|} v_{i} * w_{i},
\end{equation}
where $V$ is obtained by concatenating $V_x$ and $V_y$, $v_i$ is the component of $V$, $|V|$ the length of $V$, $U$ is learnable parameters of a fully-connected layer. We apply a simplified attention strategy to acquire the attention score $s_i$ of the feature $v_i$. 

\subsection{Global Tracing Decoder}
We adopt a 3-layer LSTM as the decoder to predict the trajectory points sequentially. In particular, the decoder uses the position and the pen tip state at time step $i-1$ to predict those at time step $i$, similar to \cite{bhunia2018handwriting,nguyen2021online}. 

During decoding, previous trajectory recovery methods \cite{bhunia2018handwriting,nguyen2021online} only utilize the initial character coding. 
As a result, the forgetting phenomenon of RNN \cite{hochreiter1997long} causes the so-called trajectory-point position drifting problem during the subsequent decoding steps, especially for characters with long trajectories. To alleviate this drifting problem, we propose a global tracing mechanism by using the glyph parsing feature $Z$ at each decoding step. The whole decoding process is shown in Fig.\ \ref{figure4-2}.

\subsection{Optimization}
\label{DefinitionOfTraj}
Similar to \cite{bhunia2018handwriting,nguyen2021online}, we use the $L_1$ regression loss and the cross-entropy loss to optimize the coordinates and the pen tip states of the trajectory points, respectively. Similar to \cite{ISI:000426687100006}, during the process of optimizing pen tip states, we define  three states ``pen-down", ``pen-up" and ``end-of-sequence" respectively, which are denoted as $s_{i}^{1}, s_{i}^{2}, s_{i}^{3}$ of $p_{i}$. It is obvious that ``pen-down" data points are much more than the other two classes. To solve the biased dataset issue, we add weights (``pen-down" is set to $1$,  ``pen-up" $5$, and ``end-of-sequence" $1$, respectively) to  the cross-entropy loss. 

These hard-losses are insufficient because they require a one-to-one stroke-point correspondence, which is too strict for handwriting trajectories of variable lengths. 
We borrow the soft dynamic time warping loss (SDTW Loss) \cite{cuturi2017soft}, which has never been used for trajectory recovery, to supplement the global-alignment goal of the whole trajectory and to alleviate the alignment learning problem. 

\begin{figure*}[t]
\begin{center}
\includegraphics[width=1\linewidth]{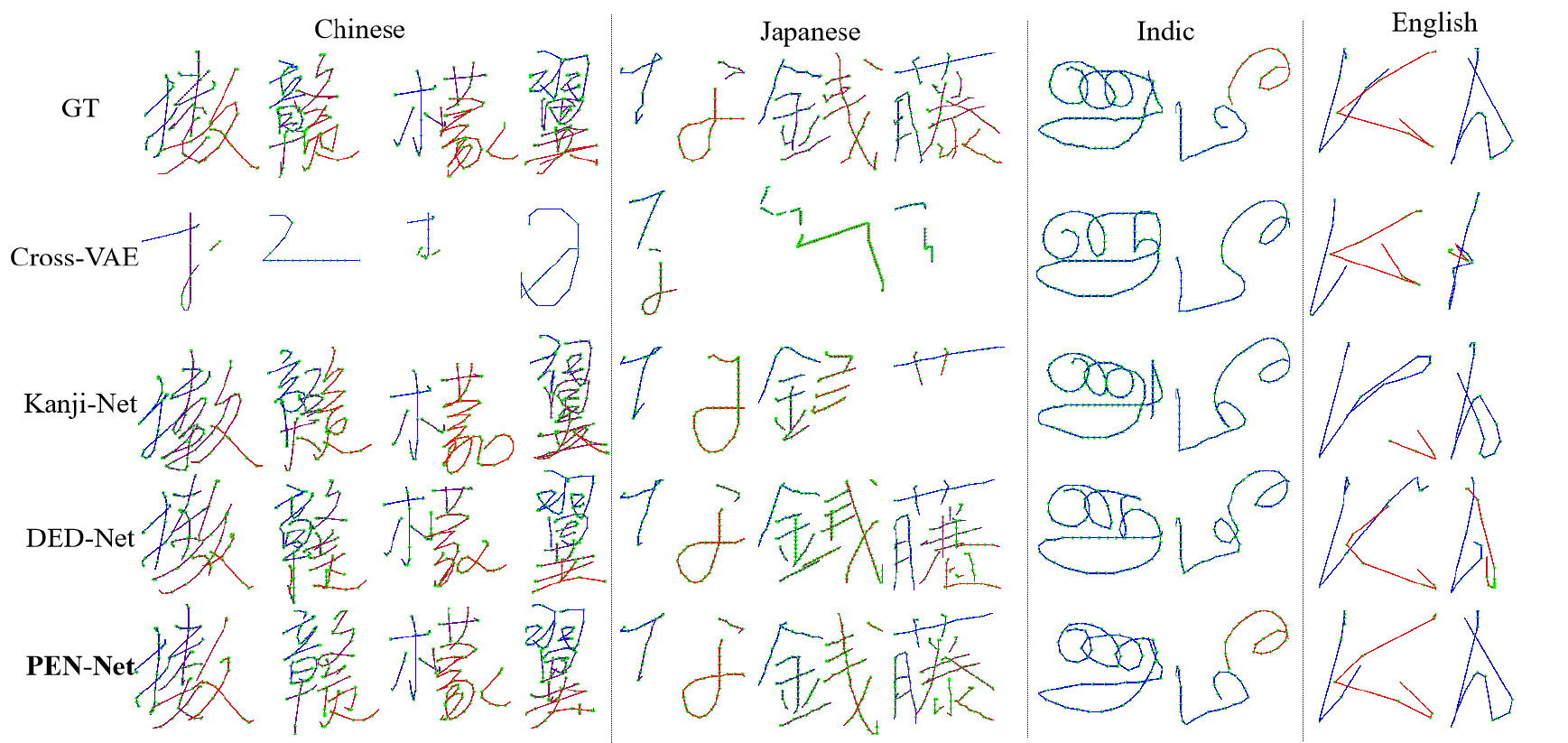}
\end{center}
  \caption{Sample visualization of recovered trajectories of our proposed PEN-Net, Cross-VAE \cite{sumi2019modality}, Kanji-Net \cite{nguyen2021online} and DED-Net \cite{bhunia2018handwriting}. Each color represents a stroke, and colors of strokes from starting to ending is represented from blue to red.}
\label{Visualization}
\end{figure*}

The DTW algorithm can solve the alignment issue during optimization using an elastic matching mechanism. 
However, since containing the hard minimization operation which is not differentiable, DTW cannot be used as an loss function directly. Hence, we place the minimization operation by a soft-minimization 
 $\min ^{\gamma}\left\{a_{1}, \ldots, a_{n}\right\}=\gamma \log \sum_{i=1}^{n} e^{-a_{i} / \gamma},  \gamma>0.
$
We define the SDTW loss
\begin{equation*}
\label{eq26}
L_{sdtw}=\operatorname{SDTW}(q, p)=\min^{\gamma}_{\phi}\left\{\sum_{t=1}^{T} d\left(q_{i_t}, p_{j_t}\right)\right\}.
\end{equation*}
The total loss is 
$L=\lambda_{1} L_1+\lambda_{2} L_{wce}+\lambda_{3} L_{sdtw}$, where $\lambda_{1}, \lambda_{2}, \lambda_{3}$ are parameters to balance the effects of the $L_1$ regression loss, the weighted cross-entropy loss and the SDTW loss, which are set to $0.5, 1$ and $1/6000$ in our experiments. 

\section{Experiments}

\subsection{Datasets}
\label{5.1datasets}

Information of datasets is given as follows, and statistics of them are in Appendix.

{\bf Chinese script.} CASIA-OLHWDB(1.0-1.2) \cite{liu2011casia} is a million-level online handwritten character dataset. We conduct experiments on all of the Chinese characters from OLHWDB1.1 which covers the most frequently used characters of GB2312-80. The largest amounts of trajectory points and strokes reach 283 (with an average of 61) and 29(average of 6), respectively. 

{\bf English script.} We collect all of the English samples from the symbol part of CASIA-OLHWDB(1.0-1.2), covering 52 classes of English letters.

{\bf Japanese script.} Referring to \cite{nguyen2021online}, we conduct Japanese handwriting recovery experiments on two datasets including Nakayosi\_t-98-09 for training and Kuchibue\_d-96-02 for testing. The largest amount of trajectory points and strokes reach 3544 (with an average of 111) and 35 (average of 6), respectively. 

{\bf Indic script.} Tamil dataset \cite{bhunia2018handwriting} contains samples of 156 character classes. The largest amount of trajectory points reach 1832 (average of 146).

\subsection{Experimental setting}
{\bf Implementation Details.} We normalize the online trajectories to [0, 64) range. In addition, in terms of the Japanese and Indic datasets, because their points densities are so high that points may overlap each other after the rescaling process, we remove the redundant points in the overlapping areas and then down-sample remaining trajectory points by half. We convert the online data to its offline equivalent by rendering the image using the online coordinate points. Although the rendered images are not real offline patterns, they are useful to evaluate the performance of trajectory recovery \cite{bhunia2018handwriting,nguyen2021online}. In addition, we train our model 500,000 iterations on the Chinese and Japanese datasets, and 200,000 iterations on the English and Indic dataset, with a RTX3090 GPU. The batch size is set to 512. The optimizer is Adam with the learning rate of 0.001.

\subsection{Comparison with State-of-the-art Approaches}
\label{SOTAsection}
In this section, we quantitatively evaluate the quality of trajectory, recovered by our PEN-Net and existing state-of-the-art methods including DED-Net \cite{bhunia2018handwriting}, Cross-VAE \cite{sumi2019modality} and Kanji-Net \cite{nguyen2021online}, on the above-mentioned four datasets via five different evaluation metrics of which AIoU and LDTW are proposed by us.

As Table. \ref{results} shows, our PEN-Net expresses satisfactory and superior performance compared to other approaches, with an average of 13\% to 20\% gap away from the second-best in all of the five evaluation criteria on the first four datasets. Moreover, to further validate the models’ effects for complex handwritings, we build two subsets by extracting 5\% of samples with the most strokes from the Japanese and Chinese testing set independently, where the number of strokes of each sample is over 15 and 10 corresponding to the two languages. According to the data(Chinese/Japanese complex in the table), PEN-Net still performs better than SOTA methods. Particularly, on Japanese complex set, PEN-Net expresses superior performance compared to other approaches, with an average of 27.3\% gap away from the second-best in all of the five evaluation criteria. 

As the visualization results in Fig.\ \ref{Visualization}, Cross-VAE \cite{sumi2019modality}, Kanji-Net \cite{nguyen2021online} and DED-Net \cite{bhunia2018handwriting} can recover simple characters' trajectories (English, Indic, and part of Japanese characters). However, their methods exhibit error phenomena, such as stroke duplication and trajectory deviation, in complex situations. Cross-VAE \cite{sumi2019modality} may fail at recovering trajectories of complex characters (Chinese and Japanese), and Kanji-Net \cite{nguyen2021online} cannot recover the whole trajectory of complex Japanese characters. In contrast, our PEN-Net makes accurate and reliable recovery prediction on both simple and complex characters, demonstrating an outstanding performance in terms of both visualization and quantitative metrics compared with the three prior SOTA works. 

\begin{table}[t]
\caption{Comparisons with state-of-the-art methods on four different language datasets. $\downarrow/\uparrow$ denote the smaller/larger, the better.}
\begin{center}\setlength{\tabcolsep}{0.5mm}
\small
\begin{tabular}{c|c|cc|ccc}
\hline
         \multirow{2}{*}{datasets}          & \multirow{2}{*}{Method} & \multicolumn{5}{c}{Evaluation metric}                                                                        \\
\cline{3-7}
                     &     & {\bf AIoU} $\uparrow$ & LPIPS $\downarrow$  & {\bf LDTW}$\downarrow$ & DTW $\downarrow$ & RMSE $\downarrow$ \\ \hline
\multirow{4}{*}{Chinese}  & Cross-VAE\cite{sumi2019modality}  &         0.146       &   0.402        &        13.64         &        1037.9          &      20.32      \\
                          & Kanji-Net\cite{nguyen2021online}  &         0.326       &   0.186        &         5.51         &        442.7          &       15.43       \\
                          & DED-Net\cite{bhunia2018handwriting} &     0.397       &   0.136      &        4.08          &      302.6            &      15.01     \\
                          & {\bf PEN-Net} &   {\bf0.450} & {\bf0.113}    & {\bf3.11} & {\bf233.8} & {\bf14.39}      \\ \hline
\multirow{4}{*}{English}  & Cross-VAE\cite{sumi2019modality}  &        0.238       &   0.206        &        7.43        &        176.7          &      20.39       \\
                        & Kanji-Net\cite{nguyen2021online} &          0.356       &    0.121    &        5.98          &       149.1           &       18.66    \\
                          & DED-Net\cite{bhunia2018handwriting} &       0.421        &   0.089    &        4.70          &        109.4          &      16.05        \\
                          & {\bf PEN-Net} &       {\bf0.461}       &     {\bf0.074}      &       {\bf3.21}           &      {\bf77.35}            &   {\bf15.12}        \\ \hline
\multirow{4}{*}{Indic}    & Cross-VAE\cite{sumi2019modality}  &         0.235       &   0.228        &       4.89       &        347.3         &      16.01     \\
                          & Kanji-Net\cite{nguyen2021online} &        0.340       &    0.163      &     3.04             &          234.0        &     15.65      \\
                          & DED-Net\cite{bhunia2018handwriting}  &       0.519         &   0.084     &       2.00         &        130.5         &    14.52      \\
                          & {\bf PEN-Net} &       {\bf 0.546}      &     {\bf 0.074}      &         {\bf 1.62}         &        {\bf 104.9}          &     {\bf 12.92}       \\ \hline 
\multirow{4}{*}{Japanese } & Cross-VAE\cite{sumi2019modality}  &       0.164      &   0.346        &        22.7         &        1652.2         &     38.79     \\
                          & Kanji-Net\cite{nguyen2021online} &        0.290      &    0.236     &        6.92          &       395.0           &      19.47       \\
                          & DED-Net\cite{bhunia2018handwriting} &        0.413      &  0.150        &        4.70          &      214.0            &      18.88      \\
                          & {\bf PEN-Net} &       {\bf 0.476}        &    {\bf 0.125}      &       {\bf 3.39}           &        {\bf 144.5}          &  {\bf 17.08}         \\ \hline \hline
\multirow{4}{*}{\tabincell{c}{Chinese\\(complex)}} & Cross-VAE\cite{sumi2019modality}  &        0.159      &   0.445        &         16.15         &        1816.4          &       26.24      \\
& Kanji-Net\cite{nguyen2021online}     &         0.311        &   0.218            &         5.56            &        668.4      &     15.26       \\
&DED-Net\cite{bhunia2018handwriting}         &  0.363               &   0.168            &     4.34                &        483.4       &       16.08          \\
&{\bf PEN-Net}      &     \textbf{0.411}   &    \textbf{0.143}  &      \textbf{3.58}      &    \textbf{402.9}  &      \textbf{15.61}   \\ \hline                          
\multirow{4}{*}{\tabincell{c}{Japanese\\(complex)}} & Cross-VAE\cite{sumi2019modality}  &        0.154      &   0.489        &         41.84         &        6230.3          &       60.28      \\
                        & Kanji-Net\cite{nguyen2021online}       &      0.190        &   0.435       &         9.68            &        1264.5    &     20.32       \\
                                                &DED-Net\cite{bhunia2018handwriting} &    0.341          &   0.250       &     4.35               &        536.1          &       18.31          \\
                                                &{\bf PEN-Net}  &     \textbf{0.445}   &    \textbf{0.186}  &      \textbf{2.95}    &     \textbf{344.4}      &      \textbf{16.00}   \\ \hline
                        
\end{tabular}
\end{center}
\label{results}
\end{table}

\begin{figure}[t]
\begin{center}
\includegraphics[width=0.95\linewidth]{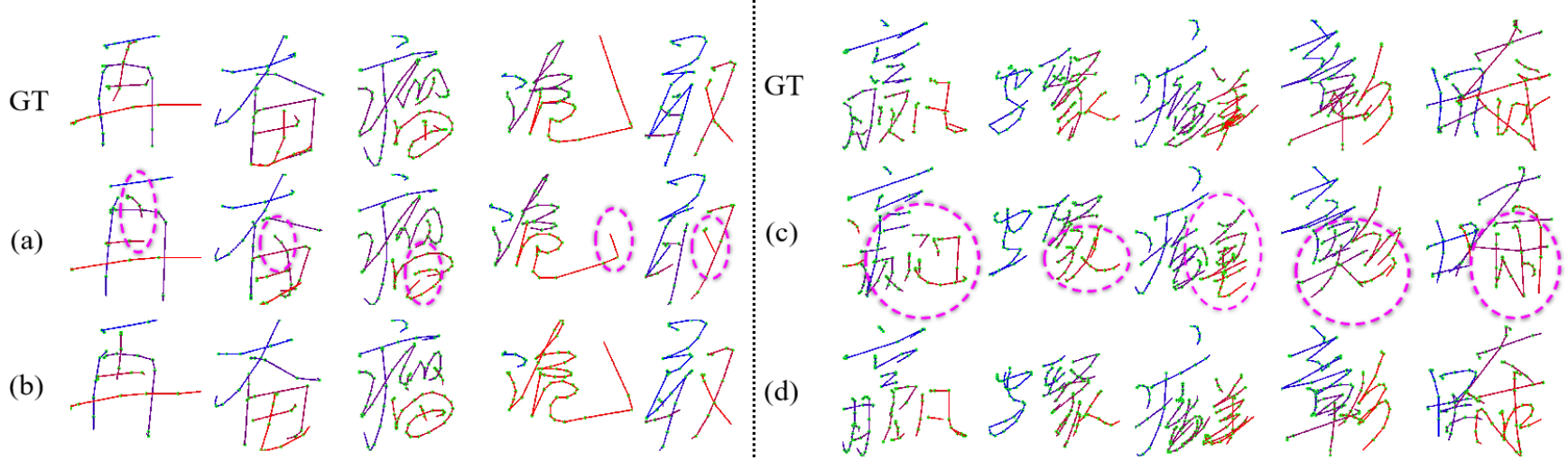}
\end{center}
   \caption{Left: Sample visualization of recovered trajectories of models (a)without and (b) with double-stream mechanism. Right: Sample visualization of models (c)without and (d) with global tracing mechanism. Stroke errors are circled in red.}
\label{figure-encoder}
\end{figure}

\begin{table}[t]
\caption{Ablation study on each component of PEN-Net.}
\begin{center}\setlength{\tabcolsep}{0.4mm}
\small
\begin{tabular}{cccc||cc|ccc}
\hline
\tabincell{c}{DS} & \tabincell{c}{GLT}  & \tabincell{c}{ATT}  & \tabincell{c}{SDTW} & {\bf AIoU}$\uparrow$ &LPIPS $\downarrow$& {\bf LDTW}$\downarrow$ & DTW$\downarrow$ & RMSE$\downarrow$  \\  \hline
       $\surd$     & $\surd$     &   $\surd$     & $\surd$      & {\bf0.451} & {\bf0.113} & {\bf3.11} & {\bf233.8} & 14.39 \\ 
        $\surd$    & $\surd$     &   $\surd$     &              & 0.433      & 0.118      & 3.53      & 261.1      &{\bf 14.06} \\ 
        $\surd$    & $\surd$     &               &              & 0.426      & 0.123      & 3.62      & 262.4      & 15.05 \\ 
        $\surd$    &             &               &              & 0.417      &  0.130     & 3.89      & 291.3      & 14.40\\ 
                   &             &               &              & 0.406      & 0.140      & 4.07      & 301.6      & 15.17  \\ \hline

\end{tabular}
\end{center}
\label{table2}
\end{table}

\subsection{Ablation Study of PEN-Net}
\label{AblationStudy}
In this section, we conduct ablation experiments on the effectiveness of PEN-Net's core components, including double-stream (DS) mechanism, global tracing (GLT) mechanism, attention (ATT) mechanism and SDTW loss. We use Chinese dataset to evaluate PEN-Net's performance for complex handwriting trajectory recovery. The evaluation metrics are the same as in Section \ref{SOTAsection}. The experiment results are reported in Table.\ \ref{table2} in which the first row relates to the full model with all components, and we gradually ablate each component one-by-one down to the plain baseline model at the bottom row. 

{\bf Double-stream mechanism.} 
In this test, we remove CRNNy from the backbone of the double-stream encoder and remains the CRNNx. As the 4th and 5th rows in Table.\ \ref{table2} show, CRNNy contributes 2.7$\%$ and 7.14$\%$ improvement on glyph fidelity metrics (AIoU and LPIPS), and 4.4$\%$, 3.41$\%$, 5.1$\%$ improvement on writing order metrics (LNDTW, DTW, RMSE). 
Additionally, as Fig.\ \ref{figure-encoder} reveals, the model without CRNNy cannot make an accurate prediction on vertical strokes of Chinese characters.

{\bf Global tracing mechanism.} As the 3rd and 4th row in Table.\ \ref{table2} show, GLT further improves the performance based on all the metrics except RMSE. The value rise in RMSE, from 14.40 to 15.05, is because this generic metric overemphasizes the point-by-point absolute deviation, which negatively affects the overall quality evaluation of the handwriting trajectory matching. In addition, as Fig.\ \ref{figure-encoder} shows, 
drifting phenomenon occurs in the recovered trajectories if GLT is removed, while, in contrast, the phenomenon disappears vise versa.

\begin{figure}[t]
\begin{center}
\includegraphics[width=1\linewidth]{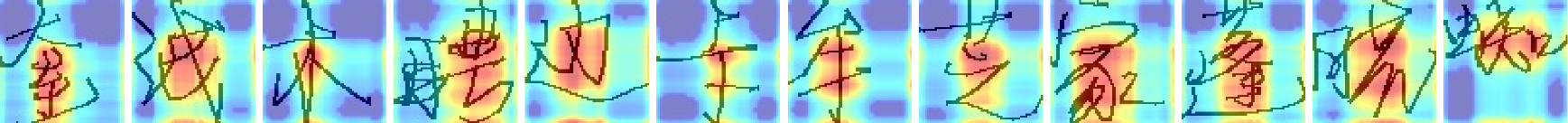}
\end{center}
   \caption{Sample visualization of attention scores maps. The maps are obtained by extracting and multiplying two attention-weighted vectors corresponding to $V_{x}$ and $V_{y}$ mentioned in Section \ref{Double-Stream Parsing Encoder}.}
\label{figure-attn}
\end{figure}

{\bf Attention mechanism.} As the 2nd and 3rd rows showed in Table. \ref{table2}, ATT also improves the performance of the model. Furthermore, as the attention heat-map visualization showed in Fig.\ \ref{figure-attn}, the stroke region always attracts more attention(in red color) than the background area(in blue color) in a character image.

{\bf SDTW loss.} 
As the 1st and the 2nd rows showed in Table.\ \ref{table2}, the SDTW loss also contributes to the performance enhancement of the model.

Finally, based on these ablation studies, the PEN-Net dramatically boost the trajectory recovery performance over the baseline by 10.8$\%$ on AIoU, 23.6$\%$ on LDTW, 22.5$\%$ on DTW, 5.1$\%$ on RMSE, 19.3$\%$ on LPIPS. Consequently, we claim that the four components of PEN-Net: double-stream mechanism, global tracing, attention mechanism and SDTW loss, all play pivotal roles w.r.t. the final performance of trajectory recovery. 

\section{Conclusion}
We have proposed two evaluation metrics AIoU and LDTW specific for trajectory recovery, and have proposed the PEN-Net for complex character recovery. 

There are several possible future directions. First, local details such as loops play an important role in some writing systems, to which we will pay more attention. Second, we have considered recovering the most natural writing order, but, as far as we know, no one has succeeded in recovering the personal writing order, which should also be a promising direction. Third, one can try to replace the decoder part by some trendy methods, e.g., transformer. Besides, we can go beyond the encoder-decoder framework, and treat this task as, for example, a decision-making problem and then use the techniques of reinforcement learning.



\subsubsection{Acknowledgements} This work has been supported by the National Natural Science Foundation of China (No.62176093, 61673182), the Key Realm Research and Development Program of Guangzhou (No.202206030001), and the GuangDong Basic and Applied Basic Rescarch Foundation (No.2021A1515012282).

%
%
%
\bibliographystyle{splncs04}
\bibliography{mybibliography}

\begin{thebibliography}{10}
\providecommand{\url}[1]{\texttt{#1}}
\providecommand{\urlprefix}{URL }
\providecommand{\doi}[1]{https://doi.org/#1}

\bibitem{al2002dynamic}
Al-Ohali, Y., Cheriet, M., Suen, C.Y.: Dynamic observations and dynamic state
  termination for off-line handwritten word recognition using hmm. In:
  Proceedings Eighth International Workshop on Frontiers in Handwriting
  Recognition. pp. 314--319. IEEE (2002)

\bibitem{archibald2021trace}
Archibald, T., Poggemann, M., Chan, A., Martinez, T.: Trace: A differentiable
  approach to line-level stroke recovery for offline handwritten text. arXiv
  preprint arXiv:2105.11559  (2021)

\bibitem{bhunia2018handwriting}
Bhunia, A.K., Bhowmick, A., Bhunia, A.K., Konwer, A., Banerjee, P., Roy, P.P.,
  Pal, U.: Handwriting trajectory recovery using end-to-end deep
  encoder-decoder network. In: 2018 24th International Conference on Pattern
  Recognition (ICPR). pp. 3639--3644. IEEE (2018)

\bibitem{boccignone1993recovering}
Boccignone, G., Chianese, A., Cordella, L.P., Marcelli, A.: Recovering dynamic
  information from static handwriting. Pattern recognition  \textbf{26}(3),
  409--418 (1993)

\bibitem{cheng2021boundary}
Cheng, B., Girshick, R., Doll{\'a}r, P., Berg, A.C., Kirillov, A.: Boundary
  iou: Improving object-centric image segmentation evaluation. In: Proceedings
  of the IEEE/CVF Conference on Computer Vision and Pattern Recognition. pp.
  15334--15342 (2021)

\bibitem{cuturi2017soft}
Cuturi, M., Blondel, M.: Soft-dtw: a differentiable loss function for
  time-series. In: International Conference on Machine Learning. pp. 894--903.
  PMLR (2017)

\bibitem{doermann1995recovery}
Doermann, D.S., Rosenfeld, A.: Recovery of temporal information from static
  images of handwriting. International Journal of Computer Vision
  \textbf{15}(1-2),  143--164 (1995)

\bibitem{guo2001forgery}
Guo, J.K., Doermann, D., Rosenfeld, A.: Forgery detection by local
  correspondence. International Journal of pattern recognition and artificial
  intelligence  \textbf{15}(04),  579--641 (2001)

\bibitem{haralick1987image}
Haralick, R.M., Sternberg, S.R., Zhuang, X.: Image analysis using mathematical
  morphology. IEEE transactions on pattern analysis and machine intelligence
  (4),  532--550 (1987)

\bibitem{hassaine2013icdar}
Hassa{\"\i}ne, A., Al~Maadeed, S., Bouridane, A.: Icdar 2013 competition on
  handwriting stroke recovery from offline data. In: 2013 12th International
  Conference on Document Analysis and Recognition. pp. 1412--1416. IEEE (2013)

\bibitem{hochreiter1997long}
Hochreiter, S., Schmidhuber, J.: Long short-term memory. Neural computation
  \textbf{9}(8),  1735--1780 (1997)

\bibitem{huang2022agtgan}
Huang, H., Yang, D., Dai, G., Han, Z., Wang, Y., Lam, K.M., Yang, F., Huang,
  S., Liu, Y., He, M.: Agtgan: Unpaired image translation for photographic
  ancient character generation. In: Proceedings of the 30th ACM International
  Conference on Multimedia. pp. 5456--5467 (2022)

\bibitem{jager1996recovering}
Jager, S.: Recovering writing traces in off-line handwriting recognition: Using
  a global optimization technique. In: Proceedings of 13th international
  conference on pattern recognition. vol.~3, pp. 150--154. IEEE (1996)

\bibitem{kato2000recovery}
Kato, Y., Yasuhara, M.: Recovery of drawing order from single-stroke
  handwriting images. IEEE Transactions on Pattern Analysis and Machine
  Intelligence  \textbf{22}(9),  938--949 (2000)

\bibitem{lallican2000off}
Lallican, P.M., Viard-Gaudin, C., Knerr, S.: From off-line to on-line
  handwriting recognition. In: Proceedings of the seventh international
  workshop on frontiers in handwriting recognition. pp. 303--312 (2000)

\bibitem{lau2005directed}
Lau, K.K., Yuen, P.C., Tang, Y.Y.: Directed connection measurement for
  evaluating reconstructed stroke sequence in handwriting images. Pattern
  Recognition  \textbf{38}(3),  323--339 (2005)

\bibitem{liu2011casia}
Liu, C.L., Yin, F., Wang, D.H., Wang, Q.F.: Casia online and offline chinese
  handwriting databases. In: 2011 International Conference on Document Analysis
  and Recognition. pp. 37--41. IEEE (2011)

\bibitem{munich2003visual}
Munich, M.E., Perona, P.: Visual identification by signature tracking. IEEE
  Transactions on Pattern Analysis and Machine Intelligence  \textbf{25}(2),
  200--217 (2003)

\bibitem{nakagawa2004collection}
Nakagawa, M., Matsumoto, K.: Collection of on-line handwritten japanese
  character pattern databases and their analyses. Document Analysis and
  Recognition  \textbf{7}(1),  69--81 (2004)

\bibitem{nel2008verification}
Nel, E.M., du~Preez, J.A., Herbst, B.M.: Verification of dynamic curves
  extracted from static handwritten scripts. Pattern recognition
  \textbf{41}(12),  3773--3785 (2008)

\bibitem{nguyen2021online}
Nguyen, H.T., Nakamura, T., Nguyen, C.T., Nakawaga, M.: Online trajectory
  recovery from offline handwritten japanese kanji characters of multiple
  strokes. In: 2020 25th International Conference on Pattern Recognition
  (ICPR). pp. 8320--8327. IEEE (2021)

\bibitem{niels2006automatic}
Niels, R., Vuurpijl, L.: Automatic trajectory extraction and validation of
  scanned handwritten characters. In: Tenth International Workshop on Frontiers
  in Handwriting Recognition. Suvisoft (2006)

\bibitem{noubigh2017survey}
Noubigh, Z., Kherallah, M.: A survey on handwriting recognition based on the
  trajectory recovery technique. In: 2017 1st International Workshop on Arabic
  Script Analysis and Recognition (ASAR). pp. 69--73. IEEE (2017)

\bibitem{otsu1979threshold}
Otsu, N.: A threshold selection method from gray-level histograms. IEEE
  transactions on systems, man, and cybernetics  \textbf{9}(1),  62--66 (1979)

\bibitem{pan1991offline}
Pan, J.C., Lee, S.: Offline tracing and representation of signatures. In:
  Proceedings. 1991 IEEE Computer Society Conference on Computer Vision and
  Pattern Recognition. pp. 679--680. IEEE (1991)

\bibitem{plamondon1999segmentation}
Plamondon, R., Privitera, C.M.: The segmentation of cursive handwriting: an
  approach based on off-line recovery of the motor-temporal information. IEEE
  Transactions on Image Processing  \textbf{8}(1),  80--91 (1999)

\bibitem{qiao2006framework}
Qiao, Y., Nishiara, M., Yasuhara, M.: A framework toward restoration of writing
  order from single-stroked handwriting image. IEEE transactions on pattern
  analysis and machine intelligence  \textbf{28}(11),  1724--1737 (2006)

\bibitem{rabhi2019handwriting}
Rabhi, B., Elbaati, A., Hamdi, Y., Alimi, A.M.: Handwriting recognition based
  on temporal order restored by the end-to-end system. In: 2019 International
  Conference on Document Analysis and Recognition (ICDAR). pp. 1231--1236. IEEE
  (2019)

\bibitem{shi2016end}
Shi, B., Bai, X., Yao, C.: An end-to-end trainable neural network for
  image-based sequence recognition and its application to scene text
  recognition. IEEE transactions on pattern analysis and machine intelligence
  \textbf{39}(11),  2298--2304 (2016)

\bibitem{sumi2019modality}
Sumi, T., Iwana, B.K., Hayashi, H., Uchida, S.: Modality conversion of
  handwritten patterns by cross variational autoencoders. In: 2019
  International Conference on Document Analysis and Recognition (ICDAR). pp.
  407--412. IEEE (2019)

\bibitem{yao2004trajectory}
Yao, F., Shao, G., Yi, J.: Trajectory generation of the writing--brush for a
  robot arm to inherit block--style chinese character calligraphy techniques.
  Advanced robotics  \textbf{18}(3),  331--356 (2004)

\bibitem{yin2016synthesizing}
Yin, H., Alves-Oliveira, P., Melo, F.S., Billard, A., Paiva, A., et~al.:
  Synthesizing robotic handwriting motion by learning from human
  demonstrations. In: Twenty-Fifth International Joint Conference on Artificial
  Intelligence (IJCAI'16). pp. 3530--3537 (2016)

\bibitem{zhang2018unreasonable}
Zhang, R., Isola, P., Efros, A.A., Shechtman, E., Wang, O.: The unreasonable
  effectiveness of deep features as a perceptual metric. In: Proceedings of the
  IEEE conference on computer vision and pattern recognition. pp. 586--595
  (2018)

\bibitem{ISI:000426687100006}
Zhang, X.Y., Yin, F., Zhang, Y.M., Liu, C.L., Bengio, Y.: {Drawing and
  Recognizing Chinese Characters with Recurrent Neural Network}. {IEEE
  TRANSACTIONS ON PATTERN ANALYSIS AND MACHINE INTELLIGENCE}
  \textbf{{40}}({4}),  {849--862} ({APR} {2018}).
  \doi{{10.1109/TPAMI.2017.2695539}}

\bibitem{zhao2020deep}
Zhao, B., Tao, J., Yang, M., Tian, Z., Fan, C., Bai, Y.: Deep imitator:
  Handwriting calligraphy imitation via deep attention networks. Pattern
  Recognition  \textbf{104},  107080 (2020)

\bibitem{zhao2018pen}
Zhao, B., Yang, M., Tao, J.: Pen tip motion prediction for handwriting drawing
  order recovery using deep neural network. In: 2018 24th International
  Conference on Pattern Recognition (ICPR). pp. 704--709. IEEE (2018)

\bibitem{zhao2019drawing}
Zhao, B., Yang, M., Tao, J.: Drawing order recovery for handwriting chinese
  characters. In: ICASSP 2019-2019 IEEE International Conference on Acoustics,
  Speech and Signal Processing (ICASSP). pp. 3227--3231. IEEE (2019)

\end{thebibliography}
%




\end{document}